%% file: main.tex
\PassOptionsToPackage{linesnumbered,ruled,vlined}{algorithm2e}
\documentclass[10pt,conference]{IEEEtran}
\IEEEoverridecommandlockouts
\usepackage{cite}
\usepackage{amsmath,amssymb,amsfonts}
\usepackage{algorithmic}
\usepackage{graphicx}
\usepackage{textcomp}

\usepackage[nolist]{acronym}
\usepackage{float}
\usepackage{fontawesome}
\usepackage{graphicx}
\usepackage{amsmath}
\usepackage{tikz}
\usepackage[ruled,vlined]{algorithm2e}
\usepackage{graphicx}
\usepackage{booktabs} 
\usepackage[linesnumbered,ruled,vlined]{algorithm2e}

\usepackage{graphicx}
\usepackage{subcaption}
\usepackage{url}
\usepackage{booktabs,tabularx}

\usepackage{tabularx}
\usepackage{makecell}

\usepackage{graphicx}

\def\BibTeX{{\rm B\kern-.05em{\sc i\kern-.025em b}\kern-.08em
    T\kern-.1667em\lower.7ex\hbox{E}\kern-.125emX}}
\begin{document}
\input{acronym}

%\title{Unleashing Low-latency Network Slicing through Adaptive UPF Path Selection in Container-Based 5G Core}
%\title{Adaptive Data Plane Path Selection for Low-Latency Network Slicing in Mobile Networks}
%\title{Slice-tailored joint path selection in B5G-Enabled Edge Computing Networks through Reinforcement Learning}
%\title{Reinforcement Learning-Based Slice-Tailored Path Selection in B5G-Enabled Edge Computing Networks}
%\title{ML-Based Coffee Leaf Disease Classification for Sustainable IoT-Driven Agriculture}
\title{Energy-Aware Ensemble Learning for Coffee Leaf Disease Classification}

\author{\IEEEauthorblockN{Larissa Ferreira {Rodrigues Moreira}\textsuperscript{1}, Rodrigo Moreira\textsuperscript{1}, Leonardo Gabriel Ferreira Rodrigues\textsuperscript{2}}
\IEEEauthorblockA{
\textsuperscript{1}Federal University of Viçosa (UFV), Minas Gerais, Brazil\\
\textsuperscript{2}Federal University of Uberlândia (UFU), Minas Gerais, Brazil\\
Emails: rodrigo@ufv.br, larissa.f.rodrigues@ufv.br, leonardo.g.rodrigues@ufu.br}
}

\maketitle

\begin{abstract}
Coffee yields are contingent on the timely and accurate diagnosis of diseases; however, assessing leaf diseases in the field presents significant challenges. Although Artificial Intelligence (AI) vision models achieve high accuracy, their adoption is hindered by the limitations of constrained devices and intermittent connectivity. This study aims to facilitate sustainable on-device diagnosis through knowledge distillation: high-capacity Convolutional Neural Networks (CNNs) trained in data centers transfer knowledge to compact CNNs through Ensemble Learning (EL). Furthermore, dense tiny pairs were integrated through simple and optimized ensembling to enhance accuracy while adhering to strict computational and energy constraints. On a curated coffee leaf dataset, distilled tiny ensembles achieved competitive with prior work with significantly reduced energy consumption and carbon footprint. This indicates that lightweight models, when properly distilled and ensembled, can provide practical diagnostic solutions for Internet of Things (IoT) applications.
\end{abstract}


\section{Introduction}\label{sec:introduction}

Coffee is among the most widely consumed and popular beverages globally. It is cultivated in diverse regions worldwide and represents a significant market. Consequently, coffee plants encounter challenges of varying severity, including diseases that affect the plant and subsequently influence the quality and productivity of the beans~\cite{Abdullah2024}. Management techniques and various technologies, particularly computational technologies, are employed to enhance productivity ~\cite{Machado2025, Bidarakundi2025}.

Among computational methodologies, \ac{AI} stands out for its ability to facilitate processes ranging from segmentation to disease classification based on crop-image diagnostics~\cite{Thakur2024}. The main challenge is not the classification itself, but rather how this capability is disseminated to end users and how they benefit from it, whether they are field professionals or agricultural producers using it for personal consumption~\cite{Holzinger2024}. This is because models based on \ac{CNNs} are generally computationally demanding, particularly during the training phase~\cite{Motta2024, Boas2025}.

While previous research has primarily concentrated on accuracy~\cite{Machado2025, Boas2025, Alkhalifa2025}, it often neglects the practical deployment challenges that influence the accessibility of image-based diagnostics for farmers to use. These challenges encompass: (i) stringent constraints on energy, memory, and intermittent connectivity linked to low-cost hardware; (ii) the necessity for robustness against variations in illumination, leaf age, cultivar differences, and camera discrepancies; (iii) issues of class imbalance and domain shift across different farms and seasons; (iv) the lack of calibrated uncertainty to support risk-aware decision-making; and (v) the limited and standardized reporting of computational and carbon costs alongside accuracy and latency~\cite{RodriguesMoreira2025, Deepak2025}.

These gaps hinder the transition from controlled benchmarks to practical applications in the field of robotics. In contrast, \ac{EL} has the potential to enhance robustness and calibration, and compress model capacity into more compact forms through knowledge transfer and distillation. This approach paves the way for on-device inference while enabling transparent reporting of accuracy and energy trade-offs, thereby promoting sustainable adoption~\cite{Delfani2024}. This is largely due to the fact that models based on \ac{CNNs} are predominantly computationally intensive, particularly during the training phase.

This study addresses this challenge by utilizing \ac{EL} techniques to enable knowledge transfer to cost-effective models that are well-suited for integration into \ac{IoT} or low-cost devices. In doing so, we contribute not only to the advancement of \ac{EL} techniques for knowledge transfer but also to a deeper understanding of the energy consumption associated with these techniques. We demonstrate that transfer learning on compact backbones, combined through three complementary \ac{EL} schemes, consistently yields accuracy gains under fixed compute budgets while maintaining a lower energy footprint.

The remainder of this paper is organized as follows: Section~\ref{sec:rw} reviews relevant prior work. Section~\ref{sec:method} presents our proposed method. Section~\ref{sec:results_and_discussion} discusses the experimental results, and Section~\ref{sec:concluding_remarks} concludes the paper with final remarks and directions for future research.

\section{Related Work}\label{sec:rw}

\begin{table*}[!ht]
\centering
\caption{Short State-of-the-Art Comparison.}
\label{tab:related_work}
\renewcommand{\arraystretch}{1.3} 
\resizebox{\textwidth}{!}{%
\begin{tabular}{@{}llcccc@{}}
\toprule
\multicolumn{1}{c}{\textbf{Approach}}     & \multicolumn{1}{c}{\textbf{Method}}                                                                                         & \textbf{Coffee focus} & \textbf{Ensemble/Fusion} & \textbf{Edge/MEC} & \textbf{TinyML/on device} \\ \midrule
Aghababaei et al.~\cite{Aghababaei2025}   & AI survey from farm to fork                                                                                                 & \faAdjust             & \faAdjust                & \faAdjust         & \faCircleO                \\ \hline
Kimutai et al.~\cite{Kimutai2024}         & AgroNet CNN with domain head and TinyML                                                                                     & \faCircle             & \faCircleO               & \faCircle         & \faCircle                 \\ \hline
Dai et al.~\cite{Dai2025}                 & Lightweight CNN/ViT with dual stream and HPO                                                                                & \faAdjust             & \faCircleO               & \faCircle         & \faAdjust                 \\ \hline
Novtahaning et al.~\cite{Novtahaning2022} & Transfer learning plus ensemble of CNNs                                                                                     & \faCircle             & \faCircle                & \faCircleO        & \faCircleO                \\ \hline
Latif et al.~\cite{Latif2025}             & Feature concatenation of CNN and handcrafted                                                                 & \faCircle             & \faCircle                & \faCircleO        & \faCircleO                \\ \hline
Guedes et al.~\cite{Guedes2024}           & Benchmark ShuffleNet versus other CNNs                                                                                      & \faCircle             & \faCircleO               & \faAdjust         & \faAdjust                 \\ \hline
Thakur et al.~\cite{Thakur2024}           & \begin{tabular}[c]{@{}l@{}}Hybrid segmentation and classification \\ with U-Net, SegNet, VGG16, Mask R-CNN\end{tabular}     & \faCircle             & \faAdjust                & \faCircleO        & \faCircleO                \\ \hline
\textbf{Ours}                             & \begin{tabular}[c]{@{}l@{}}Knowledge distillation with compact CNNs \\ and simple/optimized stacking ensembles\end{tabular} & \faCircle             & \faCircle                & \faAdjust         & \faCircle                 \\ \bottomrule
\end{tabular}
}
\end{table*}

Aghababaei et al.~\cite{Aghababaei2025} survey AI across the agro-food chain, noting sustainability gains from smarter control, edge sensing, and predictive planning; they aggregate prior results but do not target lightweight models. Kimutai et al.~\cite{Kimutai2024} evaluate domain-adaptation for pest and disease detection and show field-oriented feasibility with TinyML; effectiveness is solid on source data and moderate on new crops, including a real-farm pilot.

Dai et al.~\cite{Dai2025} study compact vision models with edge quantization, reporting very high accuracy and measured latency/power on low-cost devices, emphasizing on-device efficiency. Novtahaning et al.~\cite{Novtahaning2022} combine multiple pretrained models for coffee leaves, reaching high test accuracy but without a focus on lightweight deployment.

Latif et al.~\cite{Latif2025} fuse learned and handcrafted features with a simple classifier, achieving very high scores while keeping the feature extractor relatively light. Albuquerque and Guedes~\cite{Guedes2024} benchmark multiple networks for coffee leaves and identify a lightweight option with strong generalization suitable for Agricultural \ac{IoT}. Thakur et al.~\cite{Thakur2024} couple segmentation and classification to improve localization and accuracy; the approach is not lightweight but supports targeted treatment and potential reductions in chemical use.

In Table~\ref{tab:related_work} the column ``Method'' briefly names the main technique or model. ``Coffee focus'' indicates whether the study targets coffee leaf disease or stress. ``Ensemble or Fusion'' marks whether multiple models or feature sets are combined. ``Edge or \ac{MEC}'' signals execution or orchestration close to sensors or at the edge. ``TinyML'' or on device denotes feasibility on constrained devices such as mobile or microcontroller class hardware. Status markers are used as follows: {\scriptsize \faCircle} Yes; {\scriptsize \faCircleO} No; and {\scriptsize \faAdjust} Partial.

\textbf{Positioning.} Unlike prior work centered on accuracy or cloud-heavy models, we pursue sustainable on-device diagnosis by distilling high-capacity \ac{CNN} into compact backbones and pairing them with simple and optimized ensembles. We also report computation, energy, and carbon alongside accuracy and latency, putting deployability under tight device and connectivity constraints at the center.

\section{Proposed Approach} \label{sec:method}

We introduce an energy-aware pipeline that pairs lightweight \ac{CNN} backbones with simple ensembling (optimized averaging and stacking) to deliver high-accuracy, on-device diagnosis under tight compute, bandwidth, and power budgets.

\subsection{IoT-Driven Data Acquisition Framework} \label{sec:iot_framework}

The proposed approach is situated within an IoT-driven framework designed to facilitate sustainable agricultural practices. This paradigm relies on the timely and accurate acquisition of in-situ data to enable precision agriculture, where resources such as water and pesticides can be applied in a targeted manner. Our conceptual architecture involves a network of data capture points, such as farmer-operated mobile devices or low-cost stationary cameras, deployed within the coffee plantation.

A primary challenge in such IoT systems is the significant constraint on computational power, network bandwidth, and energy consumption. Therefore, our entire ML pipeline is designed for feasibility in this context. We prioritize computationally efficient, lightweight models that are viable for deployment on edge devices or local servers. This design allows for rapid, on-site diagnostics, reducing latency and data transmission costs. %The workflow enables farmers to receive real-time alerts from the system, allowing for immediate, targeted intervention rather than broad-spectrum treatments, thereby promoting sustainable crop management and reducing environmental impact.

\subsection{Dataset} \label{sec:dataset}

\begin{figure}[!ht]
    \centering
    \resizebox{\columnwidth}{!}{%
    \begin{tabular}{cccc}
		\includegraphics[width=0.6\textwidth]{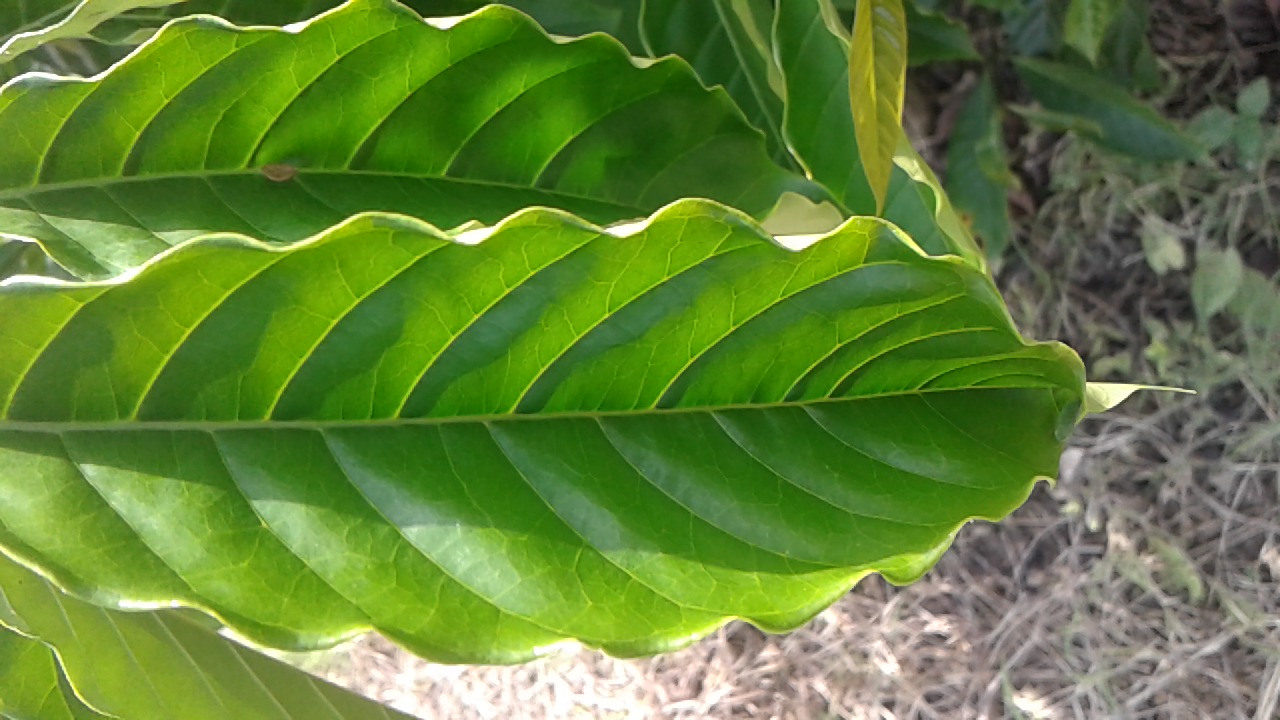} &
        \includegraphics[width=0.6\textwidth]{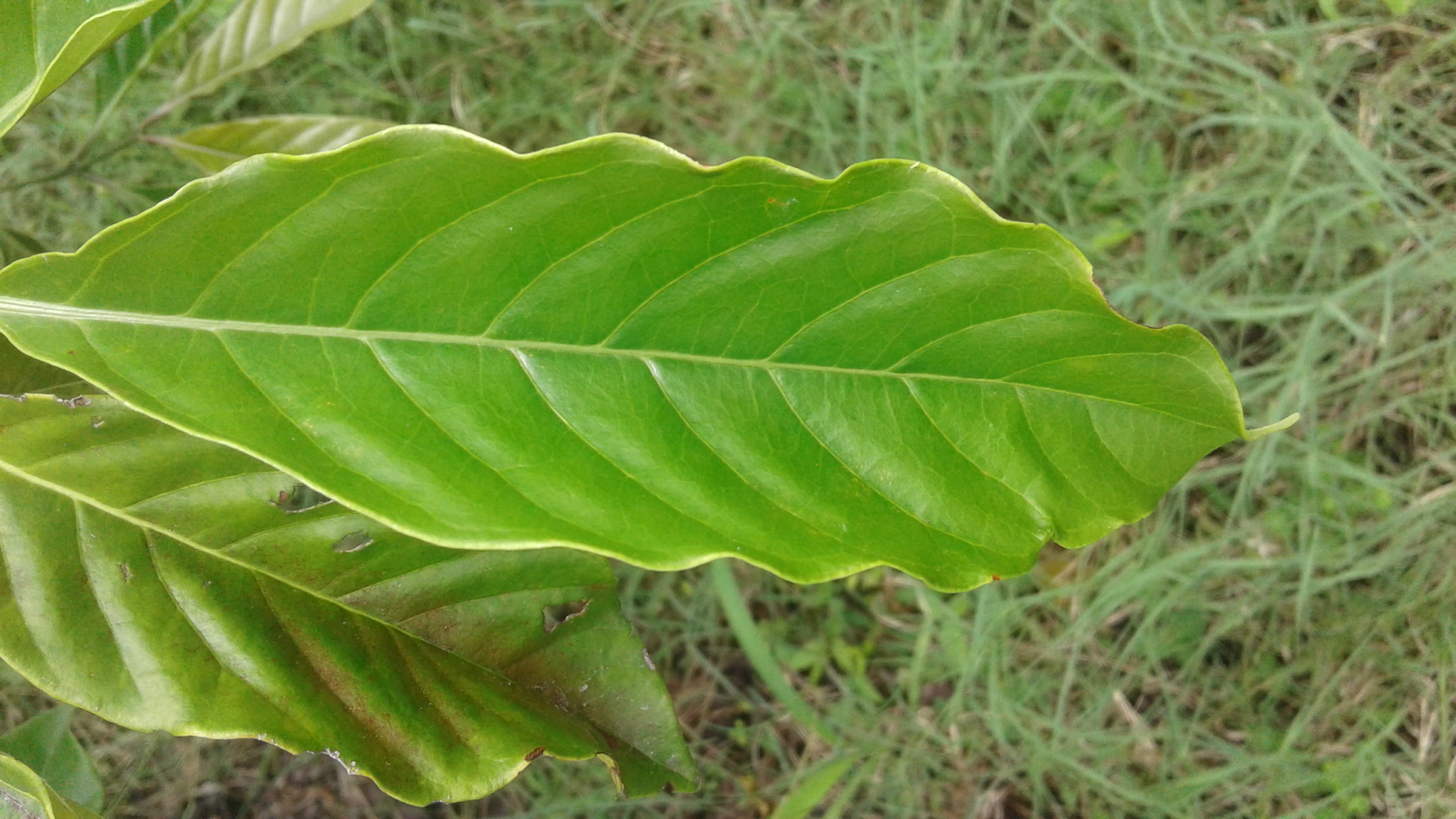} &
        \includegraphics[width=0.6\textwidth]{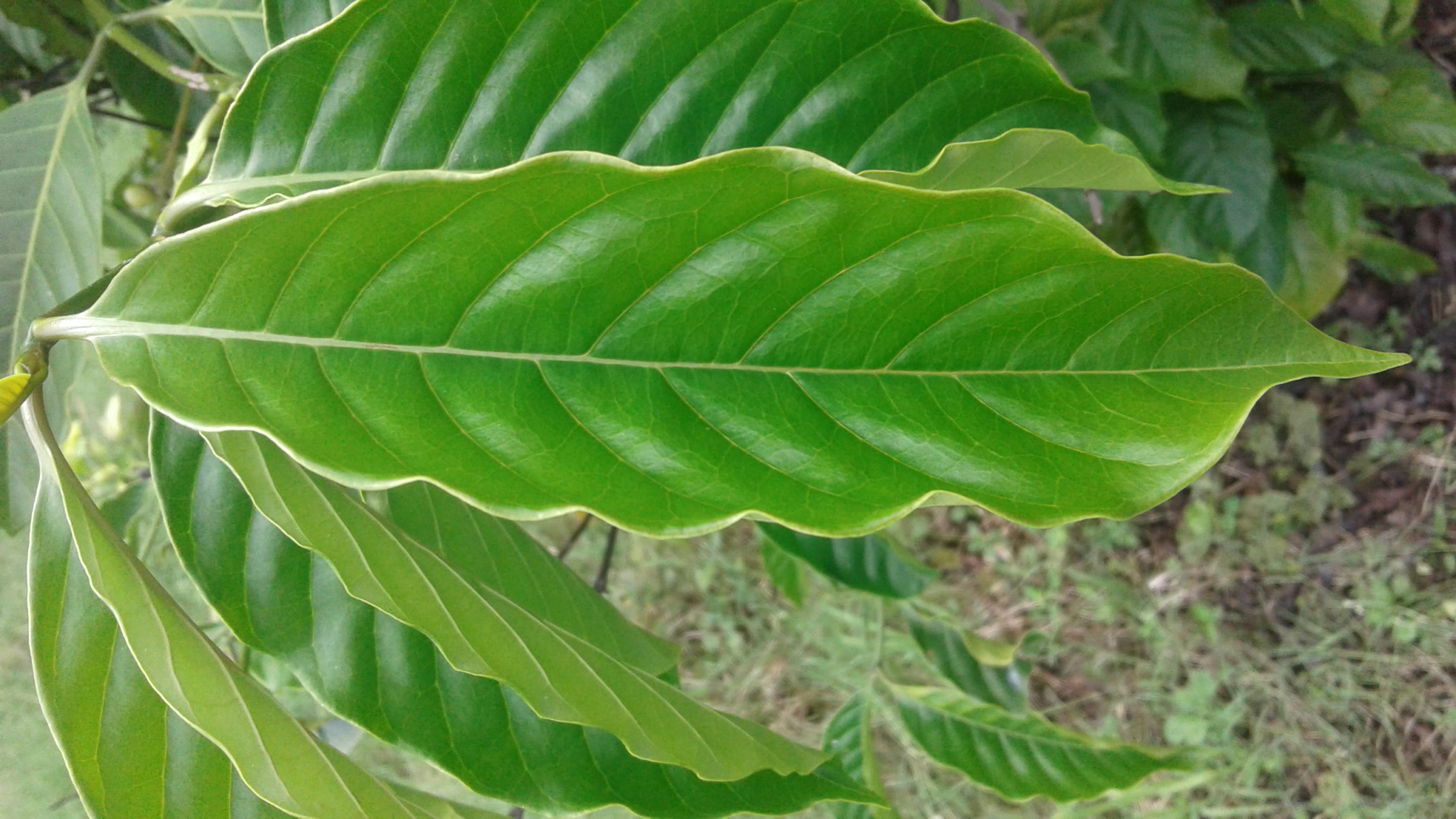} &
        \includegraphics[width=0.6\textwidth]{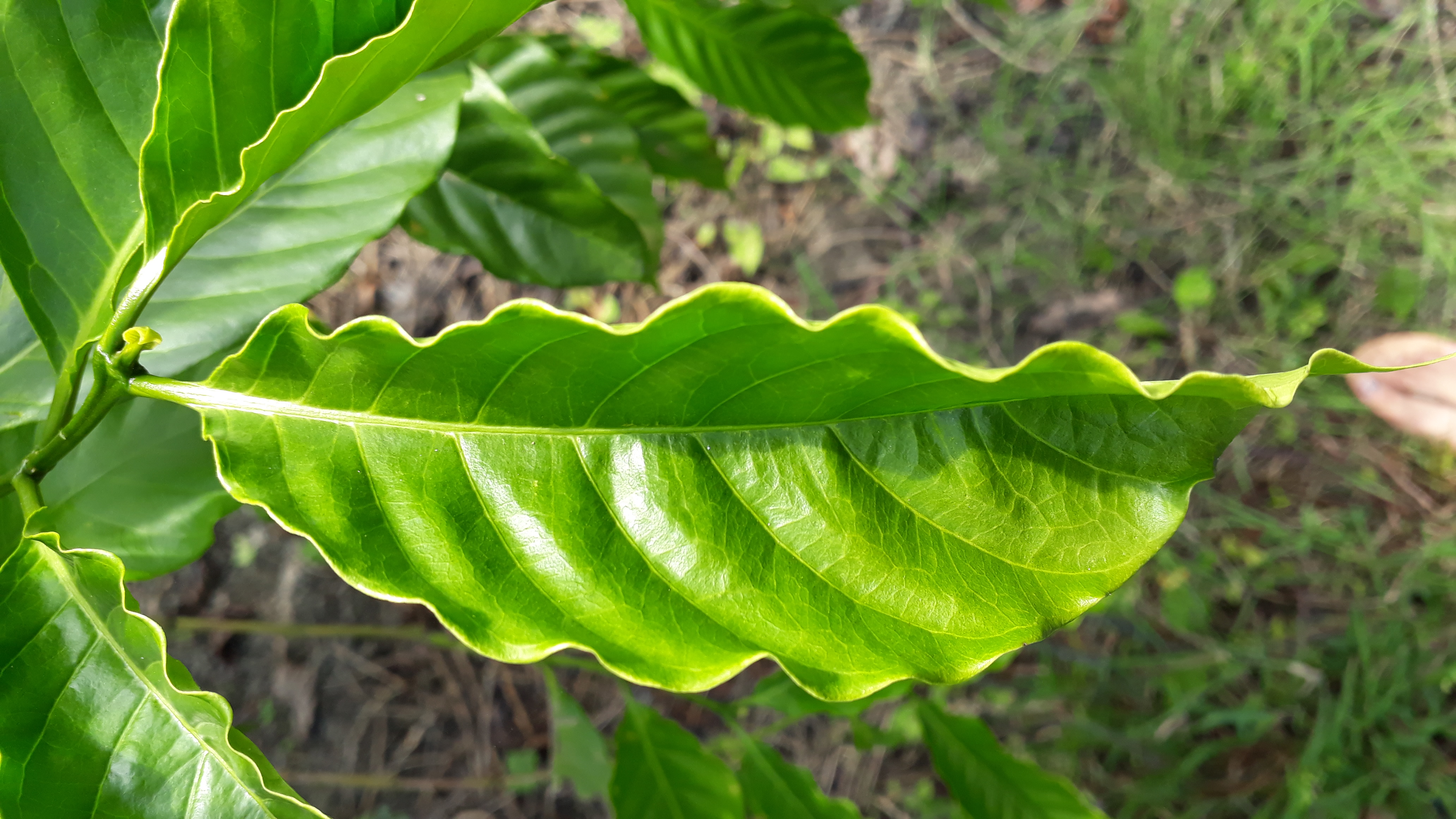} \\[0.8em]
        
		\includegraphics[width=0.6\textwidth]{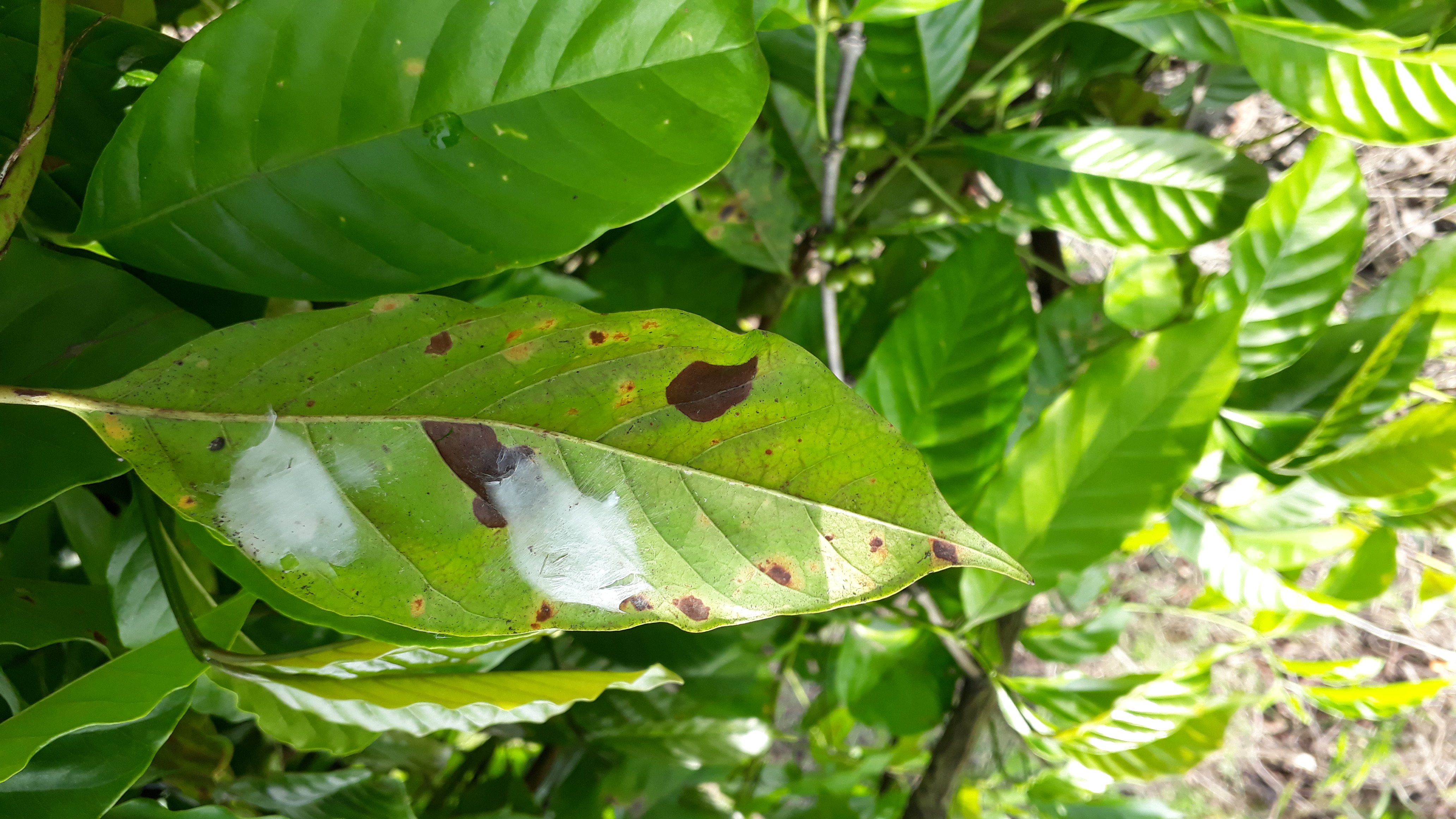} &
        \includegraphics[width=0.6\textwidth]{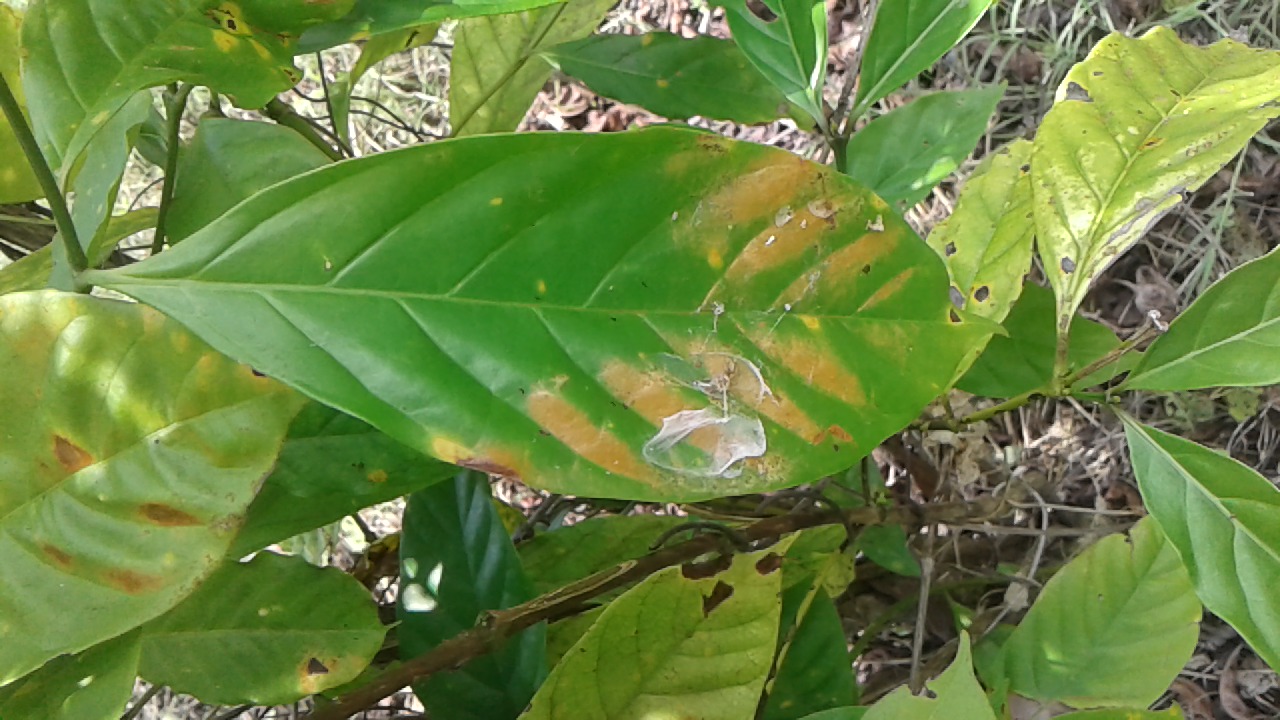} &
        \includegraphics[width=0.6\textwidth]{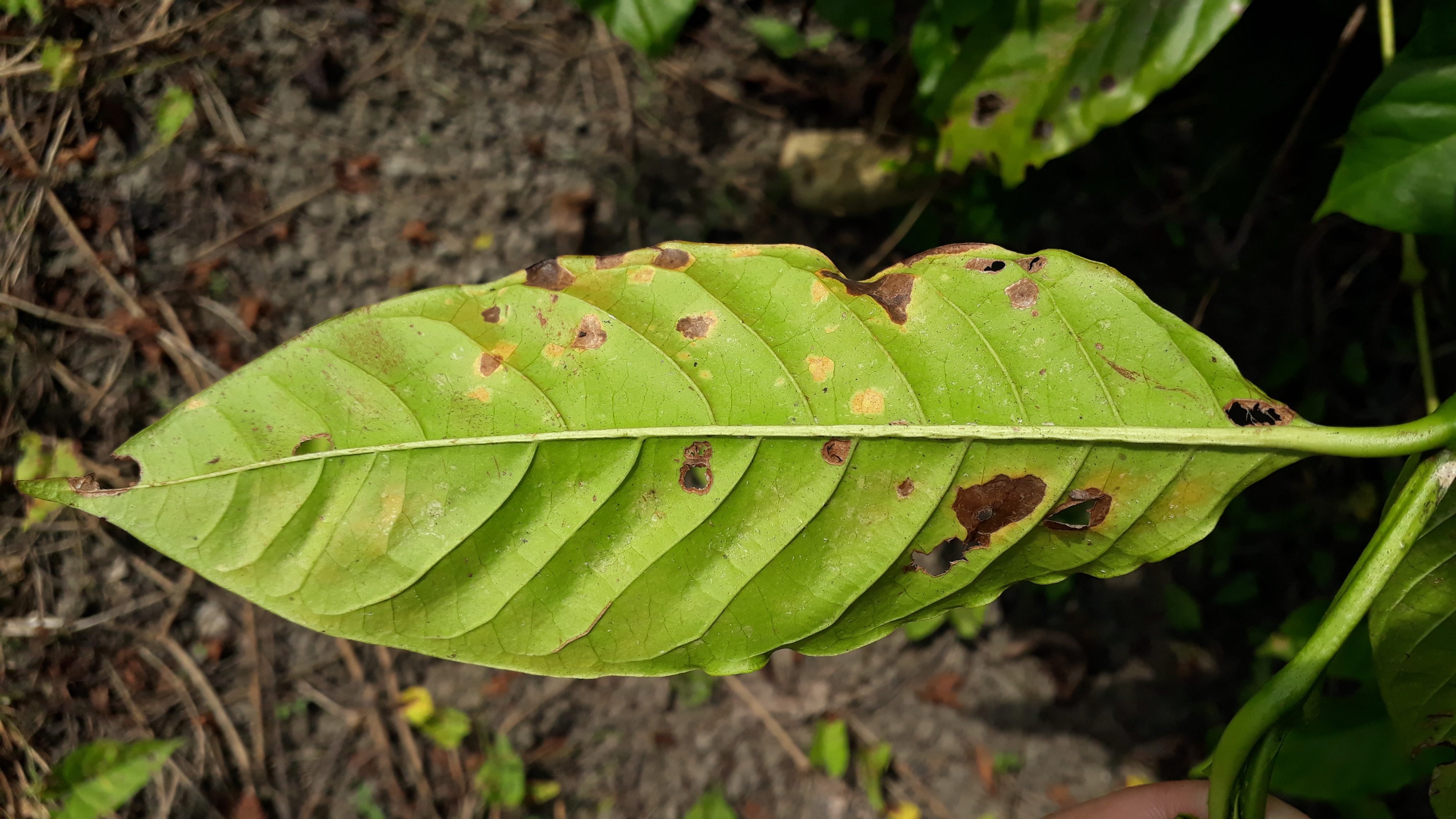} &
        \includegraphics[width=0.6\textwidth]{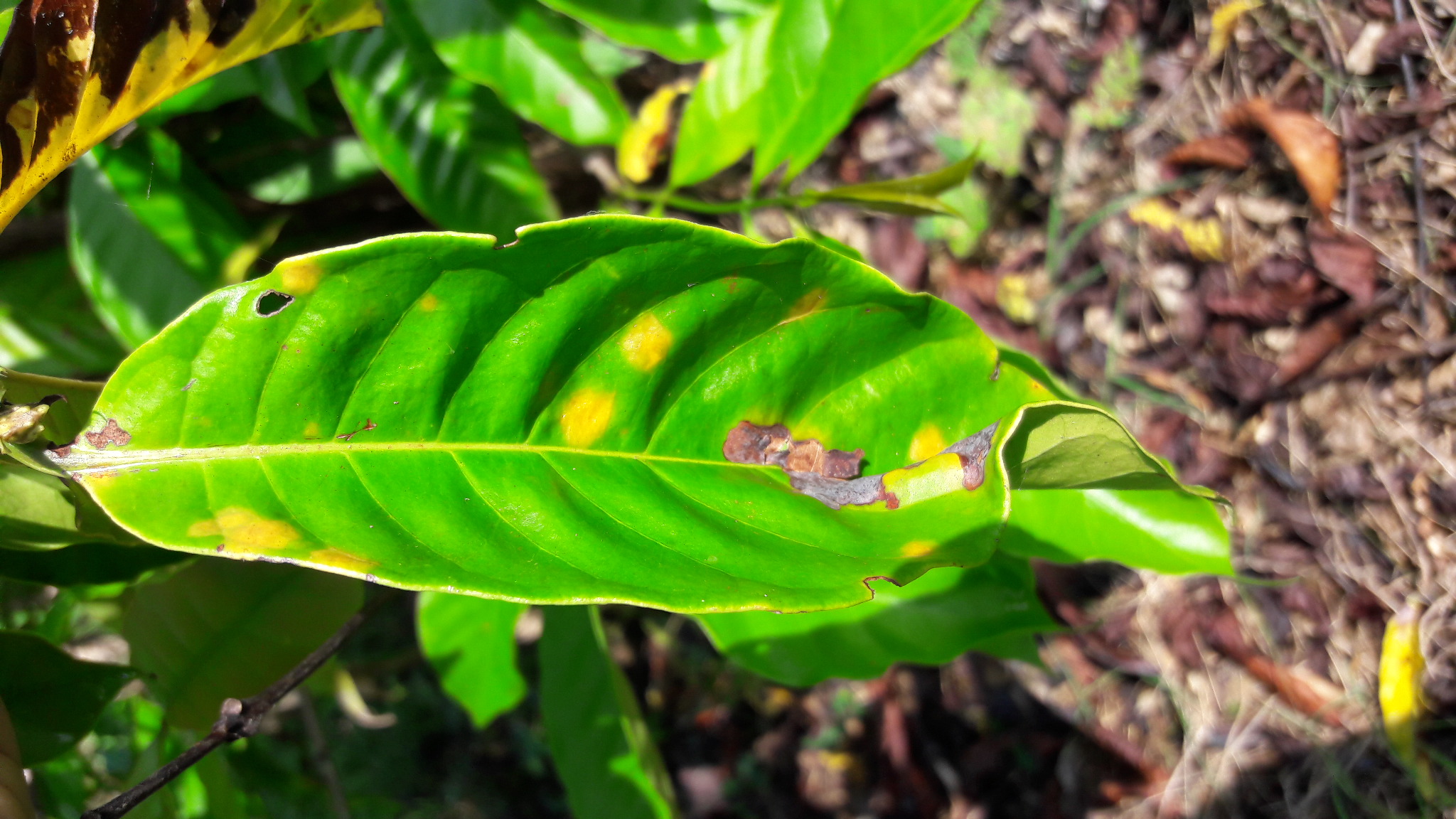} \\            
    \end{tabular}
    }
    \caption{Top row: healthy leaves. Bottom row: disease leaves with rust symptoms from the RoCoLe dataset.}
    \label{fig:exemploRocole}
    \label{fig:dataset}
\end{figure}

We used the RoCoLe coffee leaf dataset \cite{rocole} \footnote{Available at: \scriptsize{\url{https://doi.org/10.17632/c5yvn32dzg.2}}}. This dataset comprises 1560 images categorized into 2 classes with 791 healthy leaves and 769 diseased leaves including different levels of rust. To ensure a robust evaluation, the dataset was partitioned into training and test sets using a standard 80/20 stratified split. Figure~\ref{fig:dataset} shows some examples of healthy and diseased coffee leaves.

\subsection{Lightweight Base Classifiers (Base Models)} \label{sec:base_models}

We selected three \ac{CNN} architectures as our base classifiers: MobileNetV3, SqueezeNet, and ResNet-18. These models are well-established for providing a strong balance between high predictive accuracy and low computational cost, making them ideal for deployment on mobile or edge devices.

\begin{itemize}
    \item \textbf{MobileNet:} A highly efficient model designed specifically for mobile devices. It utilizes inverted residual blocks and lightweight attention (squeeze-and-excite modules) to optimize for low latency and power consumption \cite{Howard2017mobilenets}.
    
    \item \textbf{SqueezeNet:} An architecture renowned for its extremely small model size (often $<$1MB). It achieves this by aggressively reducing parameters using ``fire modules'', which consist of a 1$\times$1 \textit{squeeze} convolution layer followed by a mix of 1$\times$1 and 3$\times$3 \textit{expand} layers \cite{Iandola2016}.
    
    \item \textbf{ResNet:} An architecture that introduced ``residual connections'' (or skip connections). These connections enable the effective training of much deeper networks by mitigating the vanishing gradient problem. The 18-layer variant offers a robust baseline of high accuracy with a manageable computational load \cite{He2016CVPR}.
\end{itemize}

\subsection{Meta-Model Analysis}
\label{sec:meta_model_analysis}

The choice of the meta-model ($M_{meta}$) in the Stacking architecture is a critical design decision that directly impacts the final ensemble performance. The meta-model's task is to learn the complex relationships and error patterns from the concatenated probability vectors ($V_{train}$) supplied by the base models. To identify the optimal meta-model for our coffee leaf disease problem, we conducted a comparative analysis. We evaluated the following five classifiers, each representing a different learning paradigm \cite{Ruppert2004}:

\begin{itemize}
    \item \textbf{\ac{LR}:} A linear, probabilistic model that serves as a common and highly efficient baseline for classification.
    \item \textbf{\ac{SVM}:} A classifier that finds the optimal hyperplane to maximize the margin between classes, often using a kernel to handle non-linear data.
    \item \textbf{\ac{MLP}:} A simple feedforward neural network capable of learning complex, non-linear decision boundaries from the input probabilities.
    \item \textbf{\ac{RF}:} An ensemble method that constructs a ``forest'' of decorrelated decision trees and aggregates their votes, making it highly robust to overfitting.
    \item \textbf{LightGBM (LGBM):} A state-of-the-art gradient boosting framework that builds decision trees sequentially, where each new tree corrects the errors of the previous ones.
\end{itemize}

\subsection{Ensemble Strategies}\label{sec:ensemble_strategies}

To improve predictive performance, we investigated three ensemble strategies for combining the outputs of two base models ($M_A, M_B$) and all models: Simple Averaging, Optimized Weighted Averaging, and Stacked Generalization~\cite{Wolpert1992, Polikar2006}.

\subsubsection{Simple Averaging Ensemble}

This method, also known as soft voting, computes the mean of the probability vectors $P_A(x)$ and $P_B(x) \in \mathbb{R}^C$ from each base model. The final probability $P_{avg}(x)$ is calculated as shown in Equation~\ref{eq:simple_avg}:

\begin{equation}
P_{avg}(x) = \frac{1}{2} P_A(x) + \frac{1}{2} P_B(x)
\label{eq:simple_avg}
\end{equation}

The final prediction $\hat{y}$ is the class with the highest average probability (Equation~\ref{eq:simple_avg_pred}):
\begin{equation}
\hat{y} = \arg\max_{c \in \{1, \dots, C\}} \left( P_{avg}(x) \right)_c
\label{eq:simple_avg_pred}
\end{equation}

\subsubsection{Optimized Weighted Averaging Ensemble}

This method refines the simple average by assigning learnable weights, $w_A$ and $w_B$, to each model's output, as defined in Equation~\ref{eq:weighted_avg}.
\begin{equation}
P_{weighted}(x) = w_A P_A(x) + w_B P_B(x)
\label{eq:weighted_avg}
\end{equation}
The optimal weights $(w_A^*, w_B^*)$ are found by numerically maximizing the ensemble's classification accuracy on the test set, $D_{val}$. This optimization is performed subject to the constraints that the weights must be non-negative and sum to unity, as shown in Equation~\ref{eq:weights_constraint}.
\begin{equation}
w_A + w_B = 1; \quad w_A, w_B \ge 0
\label{eq:weights_constraint}
\end{equation}
We implemented this process by minimizing the negative accuracy using the \ac{SLSQP} algorithm \cite{Kraft1988}.

\subsubsection{Stacked Generalization (Stacking)}

Stacked Generalization (stacking) trains a second-level ``meta-model'' to learn the best combination of base model predictions. First, the probability vectors from $M_A$ and $M_B$ on the training set $D_{train}$ are concatenated to form a new feature vector $v_i$, as shown in Equation~\ref{eq:stacking_concat}.

\begin{equation}
v_i = [P_A(x_i) \oplus P_B(x_i)] = [p_{A,1}, \dots, p_{A,C}, p_{B,1}, \dots, p_{B,C}]
\label{eq:stacking_concat}
\end{equation}

This new dataset, $D_{meta} = \{(v_i, y_i)\}$, is used to train a meta-model, $M_{meta}$, for which we employed a classifier. For evaluation, predictions from the test set are processed similarly and fed to the trained $M_{meta}$ to yield the final prediction (Equation~\ref{eq:stacking_pred}).

\begin{equation}
\hat{y} = M_{meta}(v_{test}) = M_{meta}([P_A(x_{test}) \oplus P_B(x_{test})])
\label{eq:stacking_pred}
\end{equation}

\begin{figure*}[!ht]
    \centering
    \begin{tabular}{ccc}
        \resizebox{0.3\textwidth}{!}{\includegraphics{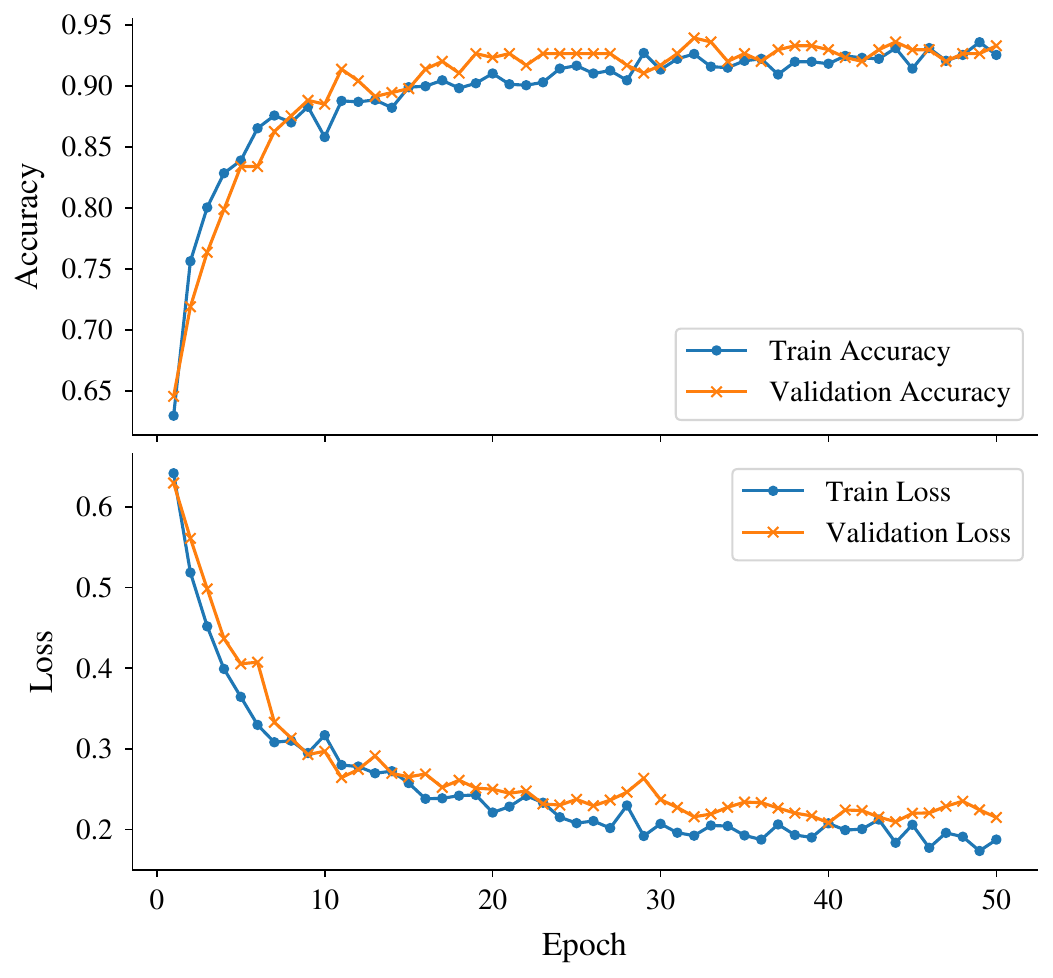}} &
        \resizebox{0.3\textwidth}{!}{\includegraphics{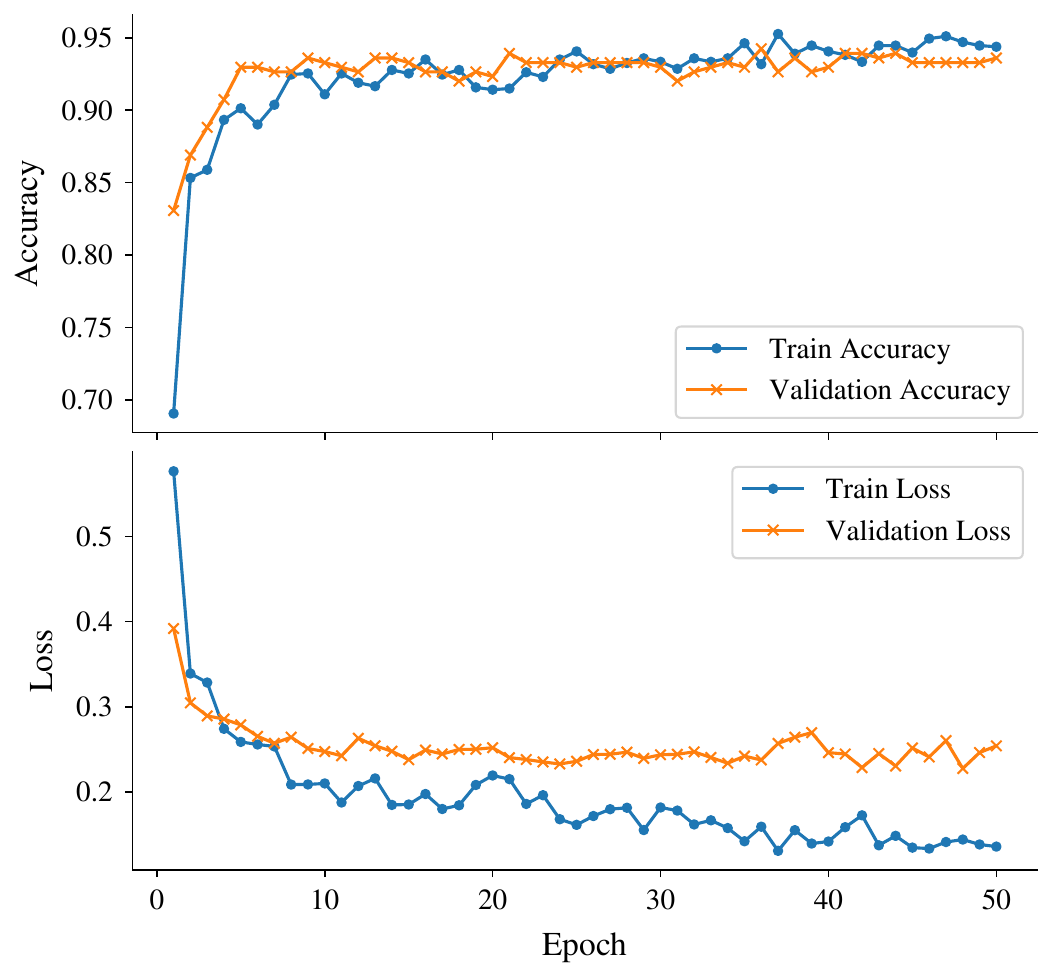}} &
        \resizebox{0.3\textwidth}{!}{\includegraphics{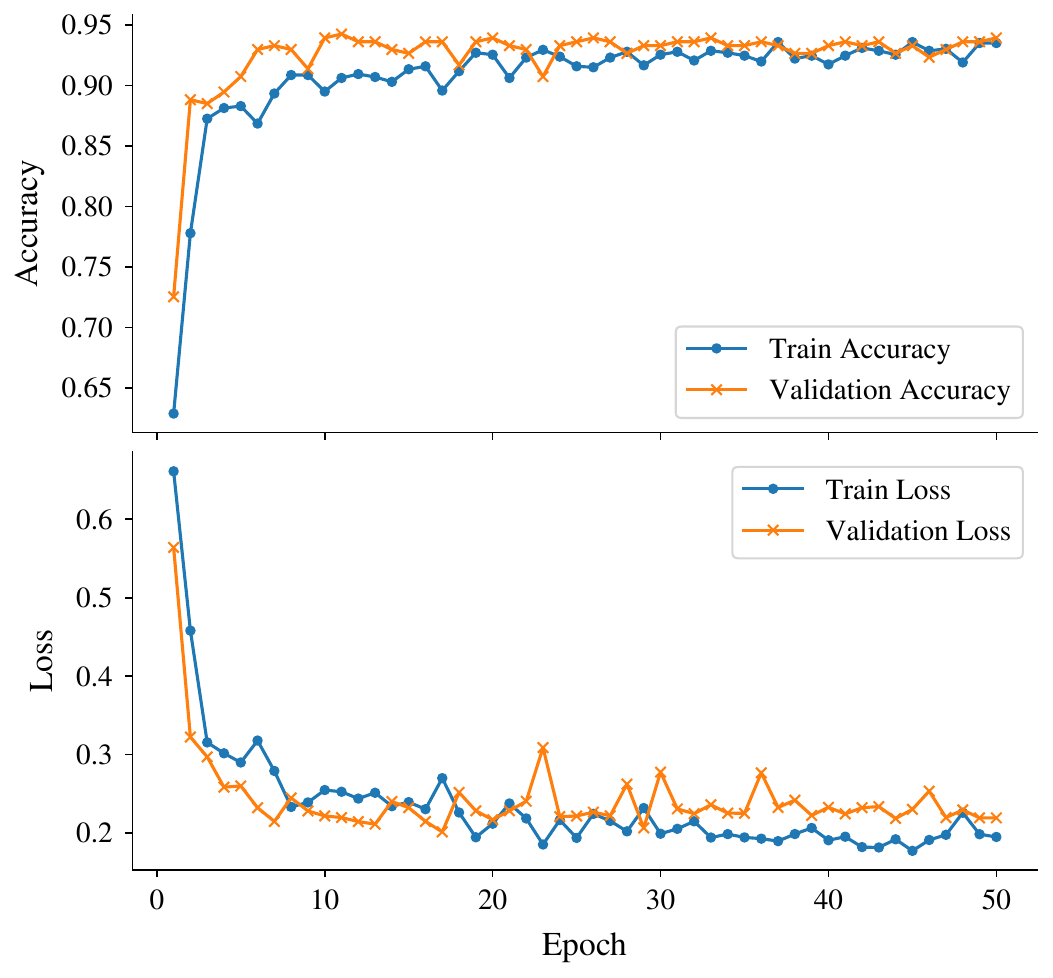}} \\
        (a) MobileNet & (b) ResNet & (c) SqueezeNet \\
    \end{tabular}
    \caption{Training and test accuracy/loss over epochs for baseline \ac{CNNs}.}
    \label{fig:train}
\end{figure*}

\section{Results and Discussion}\label{sec:results_and_discussion}

We assessed the energy consumption and carbon footprint, with experiments conducted on a cloud instance located in the Fabric testbed \cite{fabric-2019}, equipped with an NVIDIA Quadro RTX 6000 GPU and an AMD EPYC 7532 CPU. 

To enhance model generalization and prevent overfitting, the training images were subjected to a series of random transformations. This included a resize to 224 $\times$ 224 pixels, horizontal and vertical flips, and random rotation (up to 30 degrees). All images were resized and center-cropped to a final input size of 224 $\times$ 224 pixels. We employed transfer learning to leverage the feature extraction capabilities of these models. 

Each architecture was initialized with weights pre-trained on the ImageNet dataset. All base models were trained individually for 50 epochs using Stochastic Gradient Descent (SGD) with a learning rate of 0.001 and momentum of 0.9. The Cross-Entropy Loss function was adopted as the optimization criterion. First, we trained three baseline CNNs (MobileNet, ResNet, and SqueezeNet). Figure~\ref{fig:train} shows the training/test accuracy and loss curves, confirming stable convergence.

\subsection{Energy Consumption and Carbon Footprint}

The results, summarized in Table \ref{tab:energy_consumption}, highlight a clear distinction between the one-time training cost and the operational cost of the ensemble. The training phase for the three base models (MobileNet-Small, SqueezeNet, and ResNet-18) was the most resource-intensive step. All models demonstrated similar requirements, with an average training duration of 762.3 seconds ($\approx$12.7 minutes) and an average energy consumption of 0.124 kWh. 

In stark contrast, the ensemble evaluation process which includes generating predictions from two base models and training all five meta-classifiers (LR, RF, MLP, SVM, and LGBM), was exceptionally efficient. On average, each ensemble pair required only 35.1 seconds and consumed just 0.007 kWh.

\begin{table*}[ht!]
\caption{Resource consumption metrics and emissions.}
\label{tab:energy_consumption}
\resizebox{\textwidth}{!}{%
\begin{tabular}{@{}lccccccc@{}}
\toprule
\multicolumn{1}{c}{\textbf{Experiment}} & \textbf{Duration (s)} & \textbf{\begin{tabular}[c]{@{}c@{}}Energy \\ Consumed (kWh)\end{tabular}} & \textbf{\begin{tabular}[c]{@{}c@{}}CPU \\ Energy (kWh)\end{tabular}} & \textbf{\begin{tabular}[c]{@{}c@{}}GPU \\ Energy (kWh)\end{tabular}} & \textbf{\begin{tabular}[c]{@{}c@{}}RAM \\ Energy (kWh)\end{tabular}} & \textbf{\begin{tabular}[c]{@{}c@{}}Emissions \\ ($\text{kgCO}_2\text{eq}$)\end{tabular}} & \textbf{\begin{tabular}[c]{@{}c@{}}Emissions Rate \\ ($\text{kgCO}_2\text{eq/s}$)\end{tabular}} \\ \midrule
Train: Mobilenet & 
791.03 & $1.56 \times 10^{-1}$ & $1.39 \times 10^{-1}$ & $1.32 \times 10^{-2}$ & $4.20 \times 10^{-3}$ & $3.20 \times 10^{-2}$ & $4.05 \times 10^{-5}$ \\ \midrule
Train: Resnet & 
798.01 & $1.62 \times 10^{-1}$ & $1.40 \times 10^{-1}$ & $1.75 \times 10^{-2}$ & $4.30 \times 10^{-3}$ & $3.32 \times 10^{-2}$ & $4.16 \times 10^{-5}$ \\ \midrule
Train: Squeezenet & 
808.41 & $1.60 \times 10^{-1}$ & $1.42 \times 10^{-1}$ & $1.43 \times 10^{-2}$ & $4.30 \times 10^{-3}$ & $3.29 \times 10^{-2}$ & $4.07 \times 10^{-5}$ \\ \midrule
\begin{tabular}[c]{@{}l@{}}Ensemble: \\ Mobilenet + Resnet\end{tabular} & 
32.77 & $6.90 \times 10^{-3}$ & $6.20 \times 10^{-3}$ & $5.00 \times 10^{-4}$ & $2.00 \times 10^{-4}$ & $1.41 \times 10^{-3}$ & $4.31 \times 10^{-5}$ \\ \midrule
\begin{tabular}[c]{@{}l@{}}Ensemble: \\ Mobilenet + Squeezenet\end{tabular} & 
66.06 & $1.37 \times 10^{-2}$ & $1.23 \times 10^{-2}$ & $1.00 \times 10^{-3}$ & $4.00 \times 10^{-4}$ & $2.81 \times 10^{-3}$ & $4.25 \times 10^{-5}$ \\ \midrule
\begin{tabular}[c]{@{}l@{}}Ensemble: \\ Squeezenet + Resnet\end{tabular} & 
33.32 & $6.60 \times 10^{-3}$ & $5.90 \times 10^{-3}$ & $5.00 \times 10^{-4}$ & $2.00 \times 10^{-4}$ & $1.36 \times 10^{-3}$ & $4.08 \times 10^{-5}$ \\ \bottomrule
\end{tabular}%
}
\end{table*}

This shows that the energy required for the complete ensemble and stacking procedure is approximately 17.7 times less than the energy required to train a single base model. Furthermore, due to the low-carbon intensity energy grid, the estimated carbon footprint was minimal, averaging 0.437g of CO\textsubscript{2}eq for a full training run and a negligible 0.025g of CO\textsubscript{2}eq for the entire ensemble evaluation. This analysis confirms that while the initial training phase carries a moderate energy cost, the operational energy cost of using the ensemble for inference is minimal.

\begin{figure}[!ht]
    \centering
    \includegraphics[width=0.83\columnwidth]{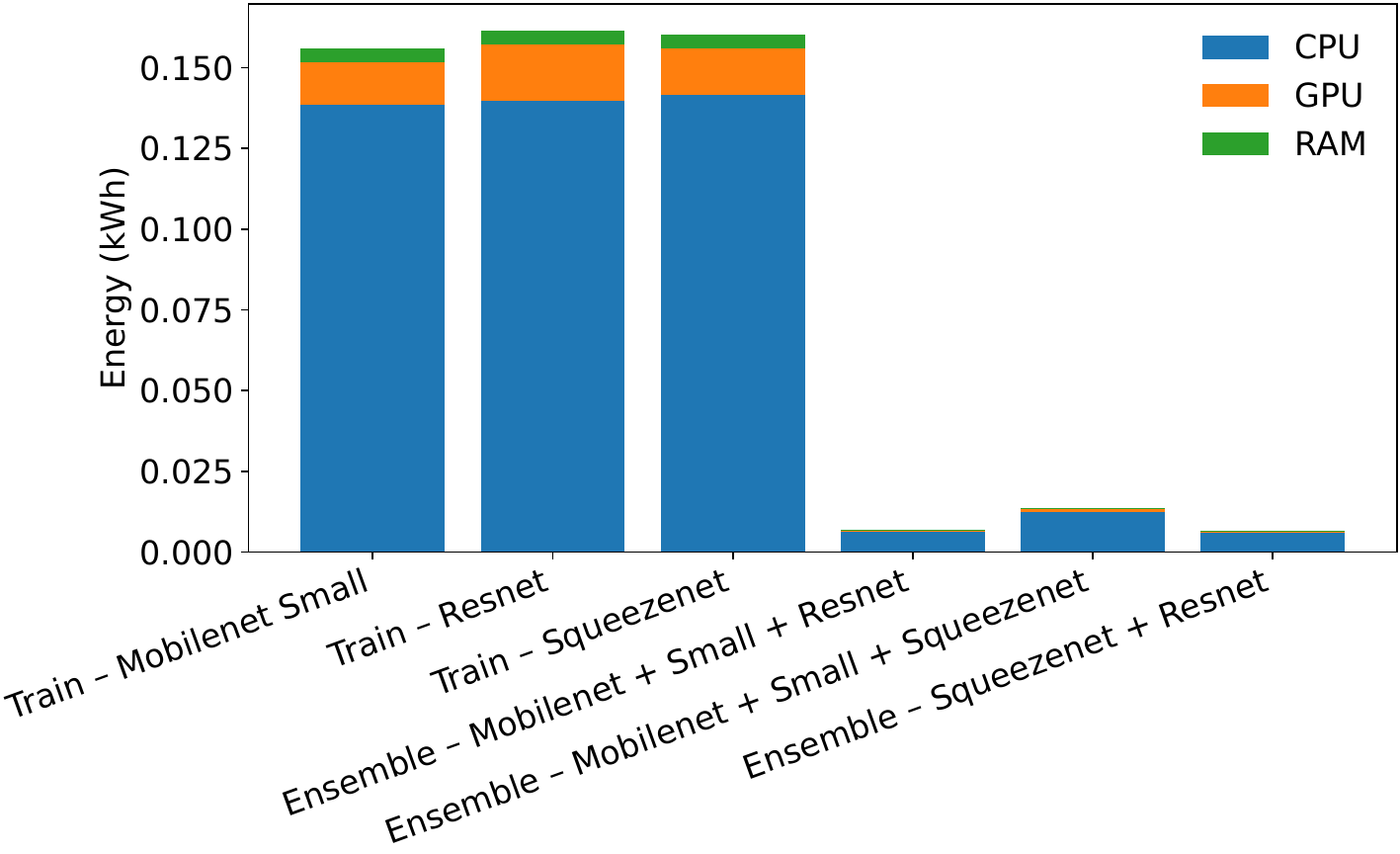}
    \caption{Energy breakdown (CPU/GPU/RAM) for training vs. two-model ensembles.}
    \label{fig:energy_breakdown}
\end{figure}

Figure \ref{fig:energy_breakdown} shows that training each baseline model (MobileNet, ResNet, SqueezeNet) dominates energy use (0.15-0.16 kWh), with consumption overwhelmingly CPU-bound; GPU and RAM contribute only marginally. In contrast, running two-model ensembles incurs over an order of magnitude less energy, remaining well below 0.02 kWh per combination. Thus, the accuracy gains from ensembling come at a negligible additional energy cost relative to training, aligning with our sustainability goals.

\subsection{Ensemble Evaluation}

The results of our meta-model analysis are summarized in Table~\ref{tab:meta_model_comparison}. This comparison reveals a high degree of performance consistency across the three evaluated classifiers. All meta-models achieved an identical accuracy of 94.25\% and recall of 94.24\%. While minor, sub-percentage variations were observed in F1-Score and Precision, the overall performance is similar, indicating that the ensemble's final accuracy is robust to the choice among these three meta-models.

\begin{table}[!ht]
\caption{Comparative performance of Stacking meta-models for the three-best model ensemble.}
\label{tab:meta_model_comparison}
\renewcommand{\arraystretch}{1.3} 
\resizebox{\columnwidth}{!}{%
\begin{tabular}{@{}lcccc@{}}
\toprule
\textbf{Meta-Model} & \textbf{Accuracy (\%)} & \textbf{F1-Score (\%)} & \textbf{Precision (\%)} & \textbf{Recall (\%)} \\ \midrule
MLP                 & 94.25                 & 94.34                 & 94.22                  & 94.24               \\
SVM                 & 94.25                 & 94.40                 & 94.21                  & 94.24               \\
LR                  & 94.25                 & 94.34                 & 94.22                  & 94.24               \\ \bottomrule
\end{tabular}
}
\end{table}

\begin{figure*}[!ht]
    \centering
    \begin{tabular}{ccc}
        \resizebox{0.31\textwidth}{!}{\includegraphics{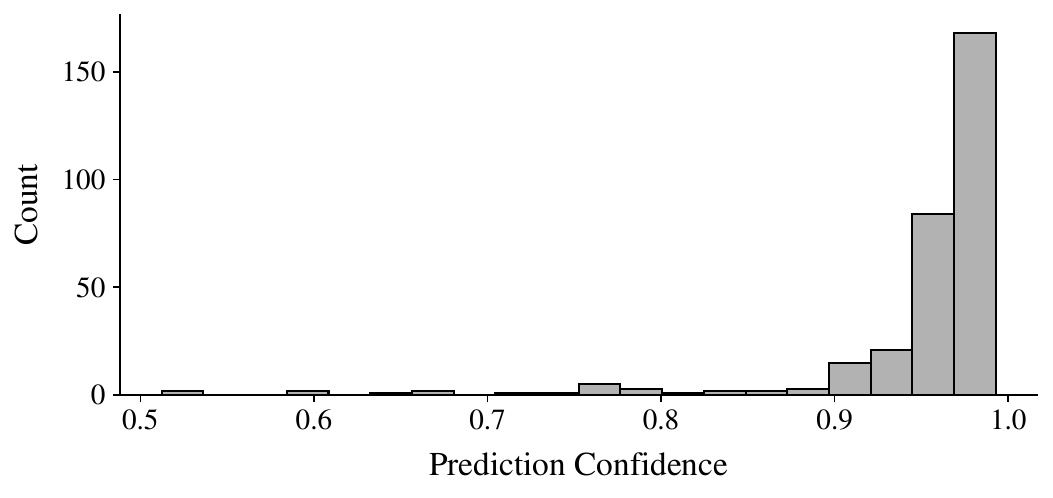}} &
        \resizebox{0.31\textwidth}{!}{\includegraphics{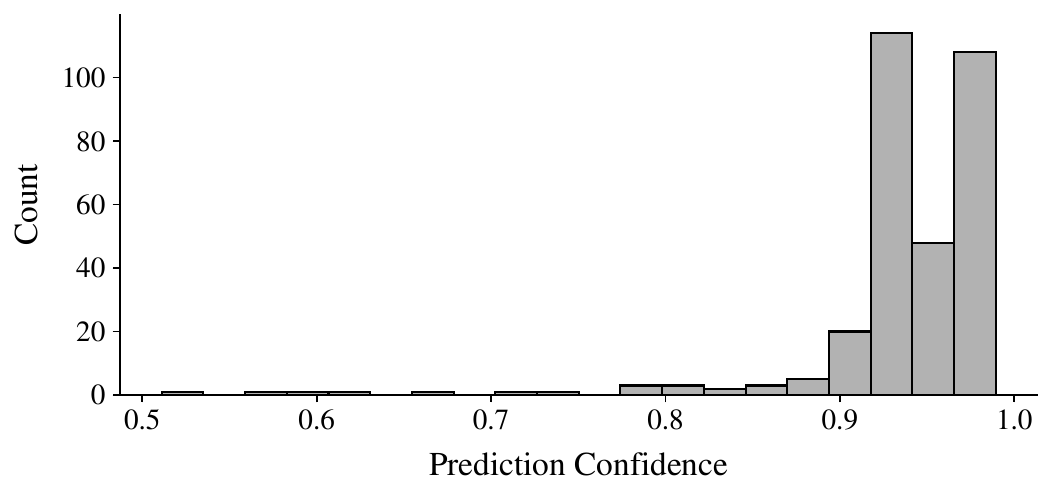}} &
        \resizebox{0.31\textwidth}{!}{\includegraphics{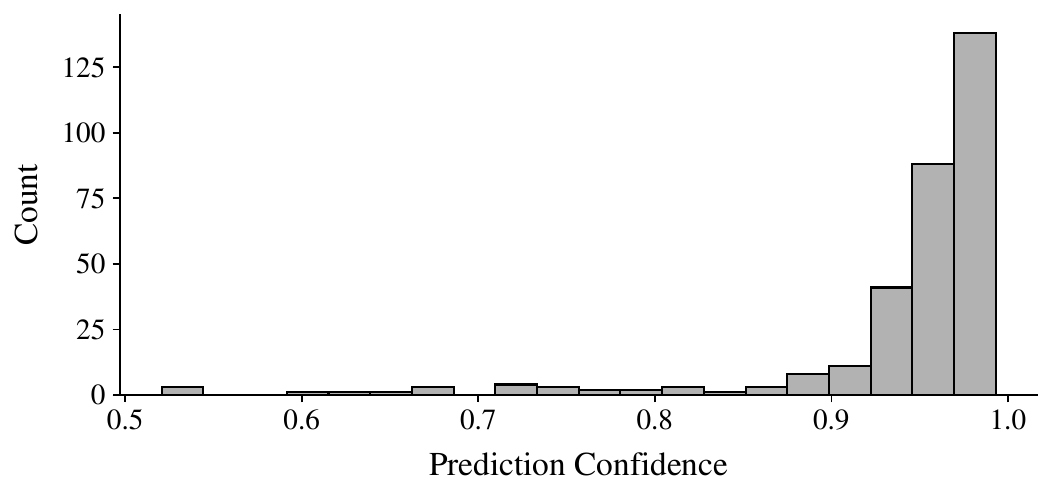}} \\
        (a) MLP & (b) SVM & (c) LR \\
    \end{tabular}
    \caption{Prediction confidence distributions for the Stacking meta-models.}
    \label{fig:confidence}
\end{figure*}

\begin{figure*}[!ht]
    \centering
    \begin{tabular}{cc}
        \resizebox{0.46\textwidth}{!}{\includegraphics{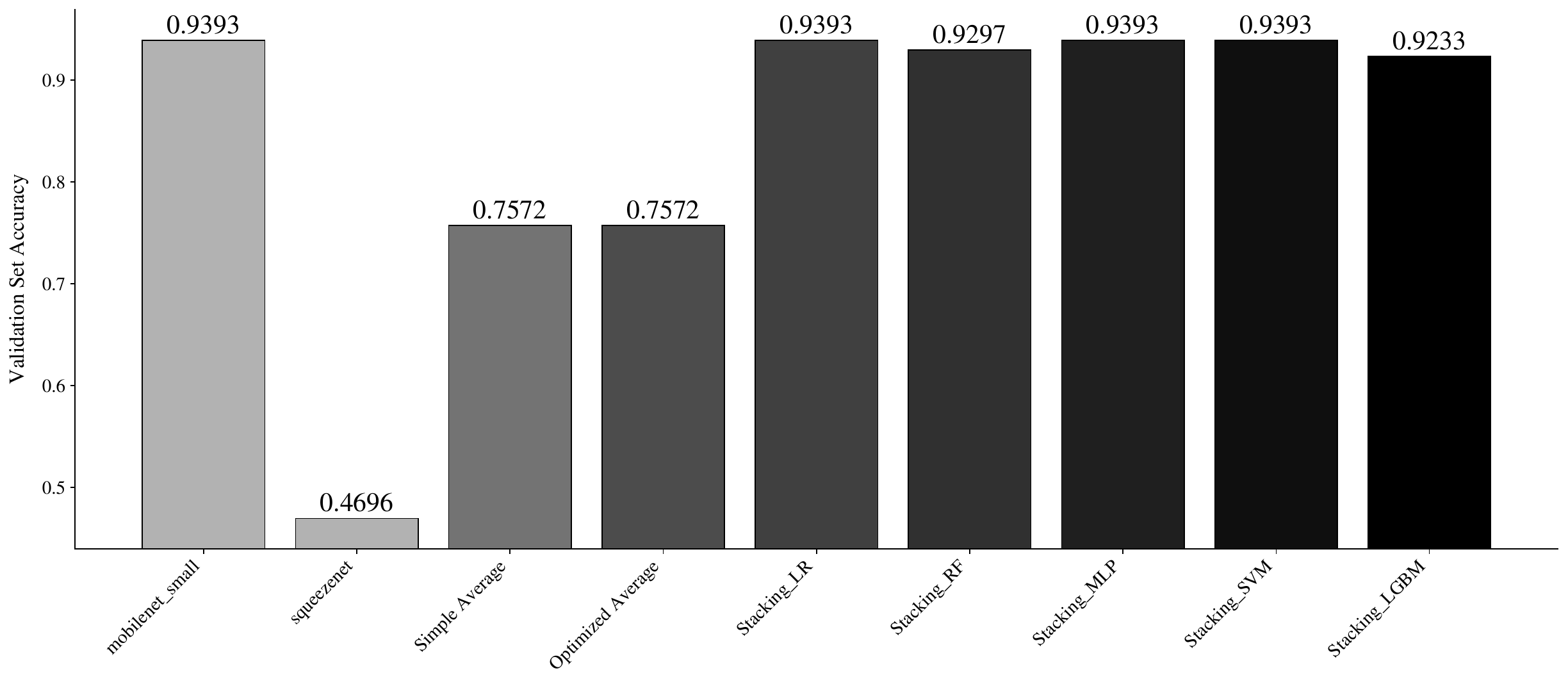}} &
        \resizebox{0.46\textwidth}{!}{\includegraphics{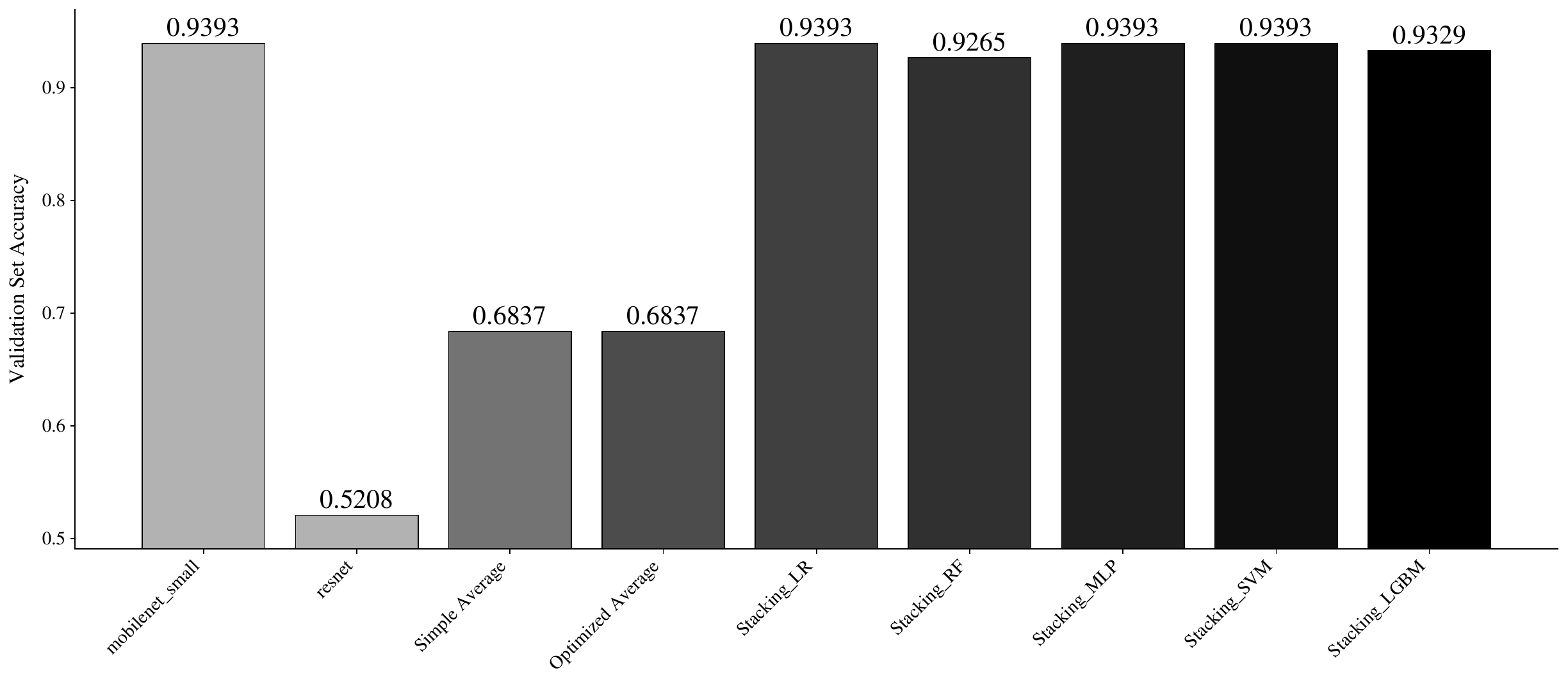}} \\
        (a) MobileNet and SqueezeNet & (b) MobileNet and ResNet \\
        \resizebox{0.46\textwidth}{!}{\includegraphics{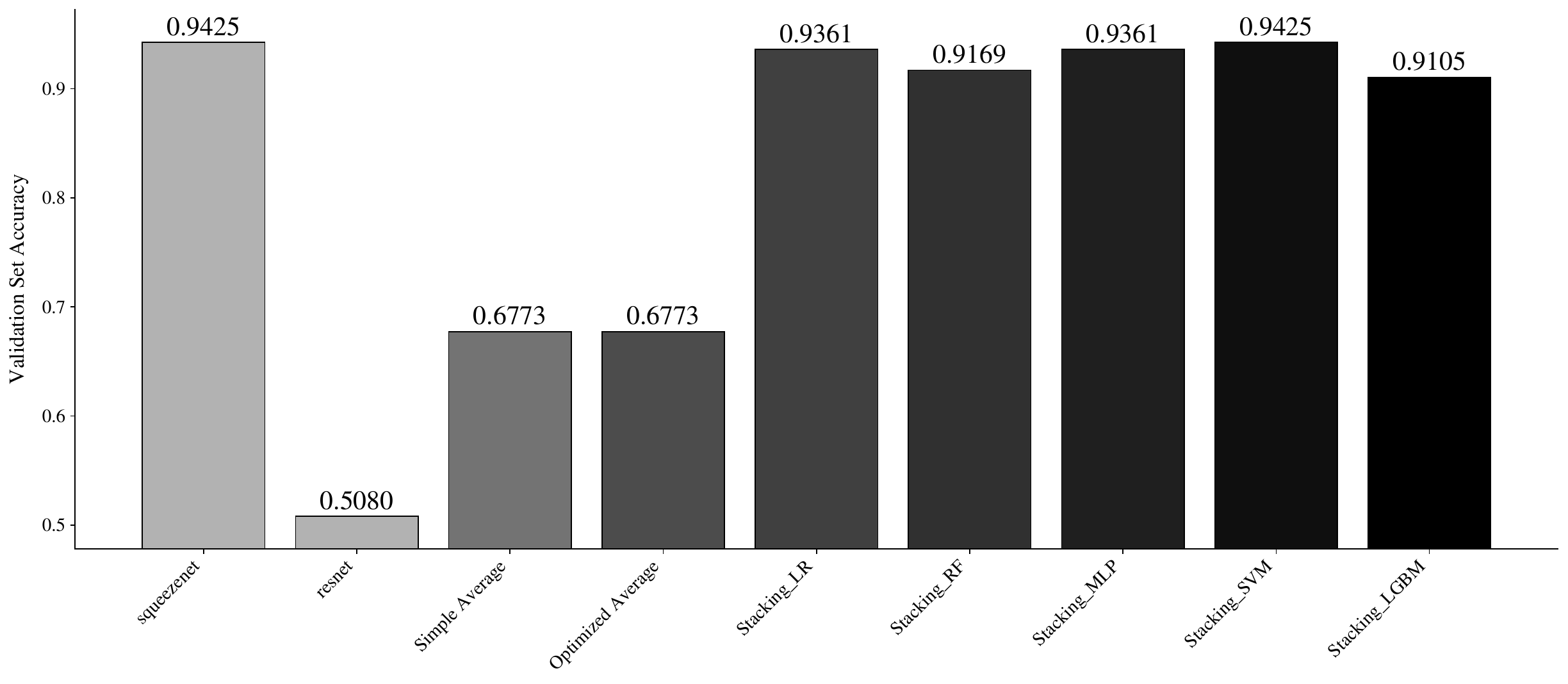}} &
        \resizebox{0.46\textwidth}{!}{\includegraphics{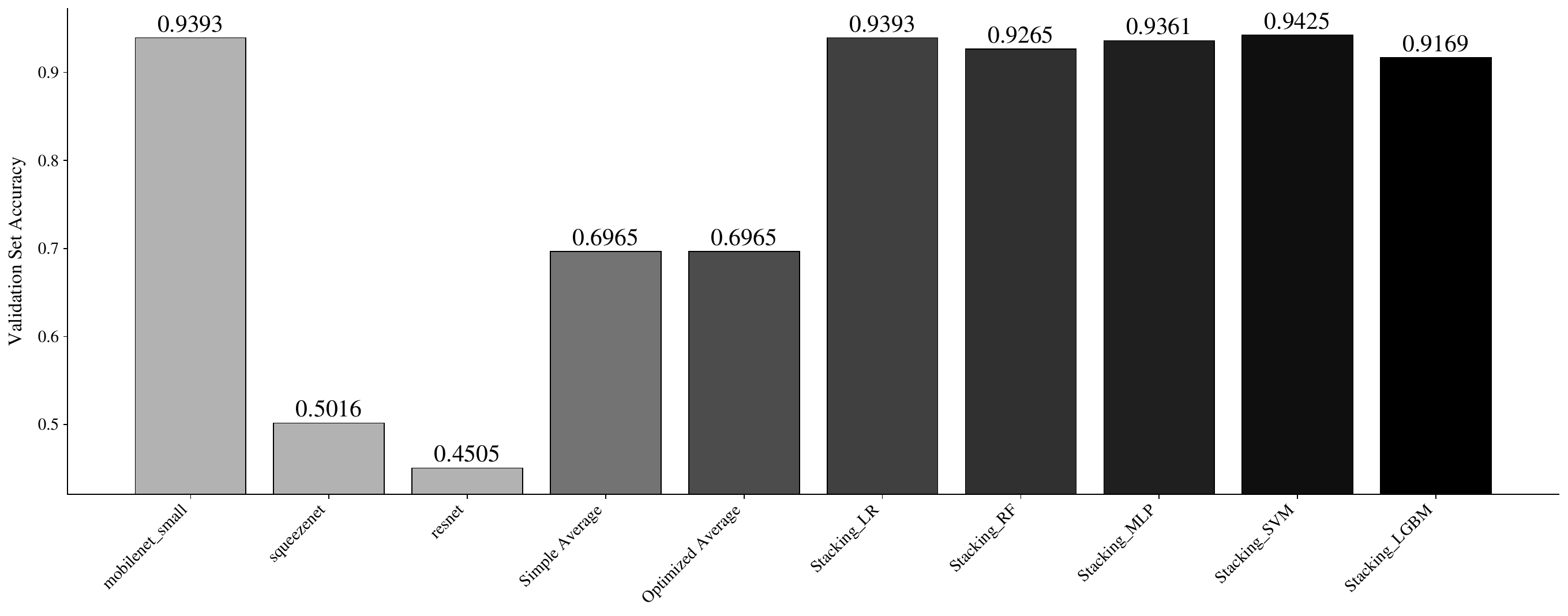}} \\
        (c) SqueezeNet and ResNet & (d) MobileNet, SqueezeNet and ResNet \\
    \end{tabular}
    \caption{Ensemble for each experiment.}
    \label{fig:ensemble}
\end{figure*}

Figure~\ref{fig:confidence} shows the prediction confidence distribution of each evaluated meta-model, the distributions is heavily left-skewed, with the vast majority of predictions falling in the 0.9 to 1.0 confidence range. This indicates the model is highly certain of its classifications, as the highest bin (confidence $>$0.95).

Prediction-confidence histograms in Fig.~\ref{fig:confidence} for the stacking meta-models concentrate near $0.95$-$1.0$, indicating highly decisive outputs. \ac{LR} shows the sharpest peak (most confident), SVM exhibits a slightly broader high-confidence mass, and \ac{MLP} has the widest spread with a small low-confidence tail. This pattern suggests well-separated classes with modest calibration differences across meta-models.

Across all pairings, we noticed in Fig~\ref{fig:ensemble} naive and optimized averaging underperform ($\approx$0.68-0.76), whereas stacking consistently recovers high accuracy ($\approx$0.93-0.94), matching or slightly exceeding the strongest single backbone. The largest relative gains arise when a strong model is paired with a weaker one (notably with ResNet), indicating that stacking effectively exploits complementary error patterns. 

\section{Concluding Remarks}\label{sec:concluding_remarks}

This study presented a knowledge distillation approach grounded in \ac{EL}, in which dense \ acn {CNNs} are trained in data centers, and their probabilities guide compact \ac{CNNs} to improve prediction quality in \ac{IoT} scenarios. Across multiple ensemble combinations, the distilled tiny \ac{CNNs} surpassed both their standalone baselines and individual dense models while operating under tight compute and energy budgets. These findings indicate that small models, when distilled and combined, can deliver accurate and efficient diagnoses on resource-constrained devices and can advance practical coffee leaf disease monitoring.

In future work, we will look at more types of \ac{CNN} backbones and new \ac{EL} strategies. We will also compare how our method holds up against Federated Learning and other training methods. We will check for accuracy, speed, data use, and energy. We plan to test how strong it is in real-world settings, study how well it makes decisions, and measure its carbon footprint during training and use. Our results help make edge intelligence better, more reliable, and of higher quality.

\section*{Acknowledgment}
The authors thank the FAPEMIG (Grant \#APQ00923-24), FAPESP MCTIC/CGI Research project 2018/23097-3 - SFI2 - Slicing Future Internet Infrastructures. 

\bibliographystyle{IEEEtran}
\bibliography{references}

\end{document}

%% file: acronym.tex
%Numbering
\acrodef{3GPP}{3rd Generation Partnership Project}
\acrodef{5G}{5th Generation Mobile Network}
\acrodef{6G}{6th Generation Mobile Network}
%-----A-----
\acrodef{AI}{Artificial Intelligence}
\acrodef{AI4Net}{\ac{AI} for Networking}
\acrodef{MDP}{Markov Decision Process}
\acrodef{AIDER}{Aerial Image Dataset for Emergency Response}
\acrodef{AMF}{Access and Mobility Management Function}
\acrodef{AIaaS}{Artificial Intelligence-as-a-Service}
\acrodef{AC}{Actor-Critic}
\acrodef{IID}{Independent and Identically Distributed}
%-----B-----
\acrodef{B5G}{Beyond Fifth Generation}
\acrodef{BPF}{Berkeley Packet Filter}
%-----C-----
\acrodef{CBR}{Constant Bit Rate}
\acrodef{CSV}{Comma-Separated Values}
\acrodef{CPU}{Central Processing Unit}
\acrodef{CNN}{Convolutional Neural Network}
\acrodef{CNNs}{Convolutional Neural Networks}
\acrodef{C-V2X}{Cellular Vehicle to-Everything}
%-----D-----
\acrodef{DoS}{Denial of Service}
\acrodef{DDQL}{Double Deep
 Q-learning}
\acrodef{DDoS}{Distributed Denial of Service}
\acrodef{DDPG}{Deep Deterministic Policy Gradient}
\acrodef{DNN}{Deep Neural Network}
\acrodef{DRL}{Deep Reinforcement Learning}
\acrodef{DQN}{Deep Q-Network}
\acrodef{DT}{Decision Tree}
\acrodef{DDQN}{Double Deep Q-Network}
\acrodef{DPDK}{Data Plane Development Kit}
%-----E-----
\acrodef{ETSI}{European Telecommunications Standards Institute}
\acrodef{eNWDAF}{Evolved Network Data Analytics Function}
\acrodef{eBPF}{Extended Berkeley Packet Filter}
\acrodef{ECDF}{Empirical Cumulative Distribution Function}
\acrodef{ECDFs}{Empirical Cumulative Distribution Functions}
\acrodef{EL}{Ensemble Learning}
%-----F-----
\acrodef{FIBRE}{Future Internet Brazilian Environment for Experimentation}
%-----G-----
\acrodef{GNN}{Graph Neural Networks}
\acrodef{GPU}{Graphics Processing Unit}
\acrodef{GPRS}{General Packet Radio Service}
\acrodef{GTP}{General Packet Radio Service Tunnelling Protocol}
\acrodef{GTP-U}{General Packet Radio Service Tunnelling Protocol User Plane}
%-----H-----
\acrodef{HTM}{Hierarchical Temporal Memory}

%-----I-----
\acrodef{IAM}{Identity And Access Management}
\acrodef{ICMP}{Internet Control Message Protocol}
\acrodef{IID}{Informally, Identically Distributed}
\acrodef{IoE}{Internet of Everything}
\acrodef{IoT}{Internet of Things}
\acrodef{ITU}{International Telecommunication Union}
\acrodef{IQR}{Interquartile Range}
\acrodef{I/O}{Input/Output}
\acrodef{IP}{Internet Protocol}
%-----J-----
%-----K-----
\acrodef{KNN}{K-Nearest Neighbors}
\acrodef{KPI}{Key Performance Indicator}
\acrodef{KPIs}{Key Performance Indicators}
%-----L-----
\acrodef{LSTM}{Long Short-Term Memory}
\acrodef{LR}{Logistic Regression}
\acrodef{LOWESS}{Locally Weighted Scatterplot Smoothing}
%-----M-----
\acrodef{MAE}{Mean Absolute Error}
\acrodef{MAD}{Median Absolute Deviation}
\acrodef{ML}{Machine Learning}
\acrodef{MLaaS}{Machine Learning as a Service}
\acrodef{MOS}{Mean Opinion Score}
\acrodef{MAPE}{Mean Absolute Percentage Error}
\acrodef{MSE}{Mean Squared Error}
\acrodef{MEC}{Multi-access Edge Computing}
\acrodef{mMTC}{Massive Machine Type Communications}
\acrodef{MFA}{Multi-factor Authentication}
\acrodef{MLP}{Multi-Layer Perceptron}
\acrodef{MADRL}{Multi-Agent Deep Reinforcement Learning}
\acrodef{MAB}{Multi-Armed Bandit}
\acrodef{MILP}{Mixed Integer Linear Programming}
%-----N-----
\acrodef{NWDAF}{Network Data Analytics Function}
\acrodef{Net4AI}{Networking for \ac{AI}}
\acrodef{NS}{Network Slicing}
\acrodef{NFV}{Network Function Virtualization}
\acrodef{NN}{Noisy Neighbor}
\acrodef{NNs}{Noisy Neighbors}
%-----O-----
\acrodef{OSM}{Open Source MANO}
\acrodef{OPEX}{Operating Expenditures}
\acrodef{O-RAN}{Open Radio Access Network}
%-----P-----
\acrodef{PCA}{Principal Component Analysis}
\acrodef{PoC}{Proof of Concept}
\acrodef{PPO}{Proximal Policy Optimization}
\acrodef{POMDP}{Partially Observable Markov decision process}
\acrodef{PCAP}{Packet Capture}
%-----Q-----
\acrodef{QoE}{Quality of experience}
\acrodef{QoS}{Quality of Service}
\acrodef{QFI}{Quality of Service Flow Identifier}
\acrodef{QFIs}{Quality of Service Flow Identifiers}
%-----R-----
\acrodef{RAM}{Random Access Memory}
\acrodef{RF}{Random Forest}
\acrodef{RL}{Reinforcement Learning}
\acrodef{RMSE}{Root Mean Square Error}
\acrodef{RNN}{Recurrent Neural Network}
\acrodef{RTT}{Round-Trip Time}
\acrodef{RAN}{Radio Access Network}
\acrodef{RTP}{Real-time Transport Protocol}
\acrodef{RIC}{RAN Intelligent Controller}
%-----S-----
\acrodef{SDN}{Software-Defined Networking}
\acrodef{SFI2}{Slicing Future Internet Infrastructures}
\acrodef{SLA}{Service-Level Agreement}
\acrodef{SON}{Self-Organizing Network}
\acrodef{SMF}{Session Management Function}
\acrodef{S-NSSAI}{Single Network Slice Selection Assistance Information}
\acrodef{SVM}{Support Vector Machine}
\acrodef{SLSQP}{Sequential Least Squares Programming}
\acrodef{SOPS}{Service-Aware Optimal
 Path Selection}
%-----T-----
\acrodef{TQFL}{Trust Deep Q-learning Federated Learning}
\acrodef{TEID}{Tunnel Endpoint Identifier}
\acrodef{TEIDs}{Tunnel Endpoint Identifiers}
%-----U-----
\acrodef{UE}{User Equipment}
\acrodef{UEs}{User Equipments}
\acrodef{UPF}{User Plane Function}
\acrodef{UPFs}{User Plane Functions}
\acrodef{PDU}{Packet Data Unit}
\acrodef{URLLC}{Ultra-Reliable and Low Latency Communications}
\acrodef{UAV}{Unmanned Aerial Vehicle}
\acrodef{UAVs}{Unmanned Aerial Vehicles}
\acrodef{UDP}{User Datagram Protocol}
%-----V-----
\acrodef{VoD}{Video on Demand}
\acrodef{VR}{Virtual Reality}
\acrodef{AR}{Augmented Reality}
\acrodef{V2V}{Vehicle-to-Vechile}
\acrodef{V2X}{Vehicle-to-Everything}
\acrodef{VNF}{Virtual Network Function}
\acrodef{VNFs}{Virtual Network Functions}

%-----W-----
%-----X-----
\acrodef{XDP}{eXpress Data Path}
%-----Y-----
%-----Z-----